\documentclass[10pt,twocolumn,letterpaper]{article}

\usepackage{cvpr}              % To produce the CAMERA-READY version
\usepackage{graphicx}
\usepackage{amsmath}
\usepackage{amssymb}
\usepackage{booktabs}
\usepackage{amsfonts,amssymb}
\usepackage[accsupp]{axessibility}
\usepackage{multirow}
\usepackage{float}
\usepackage{caption}
\usepackage{subcaption}
\usepackage{fontawesome}
\usepackage{tabularx,caption}
\usepackage[rightcaption]{sidecap}
\usepackage{xcolor}
\usepackage{colortbl}
\usepackage{psfrag}
\usepackage[percent]{overpic}
\usepackage{url}
\usepackage[accsupp]{axessibility}
\usepackage{tablefootnote}
\newcommand{\tabincell}[2]{\begin{tabular}{@{}#1@{}}#2\end{tabular}}

\def\ie{\textit{i.e., }}
\definecolor{mygray}{gray}{.9}

\usepackage[pagebackref=true,breaklinks,colorlinks,citecolor=blue,linkcolor=blue,bookmarks=false]{hyperref}

\def\xnet{CLIP2Protect\xspace}

\usepackage[capitalize]{cleveref}
\crefname{section}{Sec.}{Secs.}
\Crefname{section}{Section}{Sections}
\Crefname{table}{Table}{Tables}
\crefname{table}{Tab.}{Tabs.}

\def\cvprPaperID{9275} % *** Enter the CVPR Paper ID here
\def\confName{CVPR}
\def\confYear{2023}

\begin{document}


% \title{Privacy Preserving by User defined Adversarial Style optimized through Textual Prompts}

% \title{User-defined Textual Prompts for Naturalistic but Adversarial Style for Facial Privacy}

% \title{Facial Privacy with Style: User-defined Naturalistic "Textual" Prompts with Adversarial Effect}

% \title{Text-Driven Style for Facial Privacy Protection}

% \title{Text-Driven Robust Makeup Transfer for Facial Privacy Protection}

% \title{Text-Driven Makeup Transfer for Facial Privacy Protection using Generative Models}

% \title{Protect with Style: Text-guided Makeup for Facial Privacy Protection}

% \title{Protection with Style: Text-guided Makeup for Facial Privacy}
% \title{Protection with Style: Adversarial Latent Search via Text-guided Makeup for Facial Privacy}

%\title{CLIPGuard/CLIPShield/CLIP-Protector: Adversarial Latent Search via Text-guided Makeup for Facial Privacy}

%\title{CLIP2Protect: Adversarial Latent Search via Text-guided Makeup for Facial Privacy}
%\title{Protecting Facial Privacy using Makeup Text-Guided Adversarial Latent Search}
\title{\vspace{-0.5em}\xnet: Protecting Facial Privacy using Text-Guided Makeup via Adversarial Latent Search\vspace{-0.5em}}

\author{Fahad Shamshad \quad Muzammal Naseer \quad Karthik Nandakumar\\
Mohamed Bin Zayed University of AI, UAE\\
{\tt\small \{fahad.shamshad, muzammal.naseer, karthik.nandakumar\}@mbzuai.ac.ae}
}
%\maketitle
\iffalse
\twocolumn[{%
\renewcommand\twocolumn[1][]{#1}%
 %\vspace{-1em}
\maketitle
%\apptocmd\@maketitle{{\myfigure{}\par}}{}{}
  \centering
\vspace{-0.9cm}
  \includegraphics[width=\linewidth]{figs/title_fig.pdf}
  \captionsetup{type=figure}
  \vspace{-0.8cm}
  \captionof{figure}{}
 \vspace{0.5cm}
  \label{fig:intro_big}
 }]
 \fi
 \twocolumn[{%
\renewcommand\twocolumn[1][]{#1}%
 %\vspace{-1em}
\maketitle
 %\begin{figure*}[h]
    %\centering
    %\begin{subfigure*}%[b]{\linewidth}        %% or \columnwidth
        %\centering
        
        %\begin{overpic}[width=\textwidth,tics=10]{figs/tit_fig.pdf}
        \begin{overpic}[width=\textwidth,tics=10]{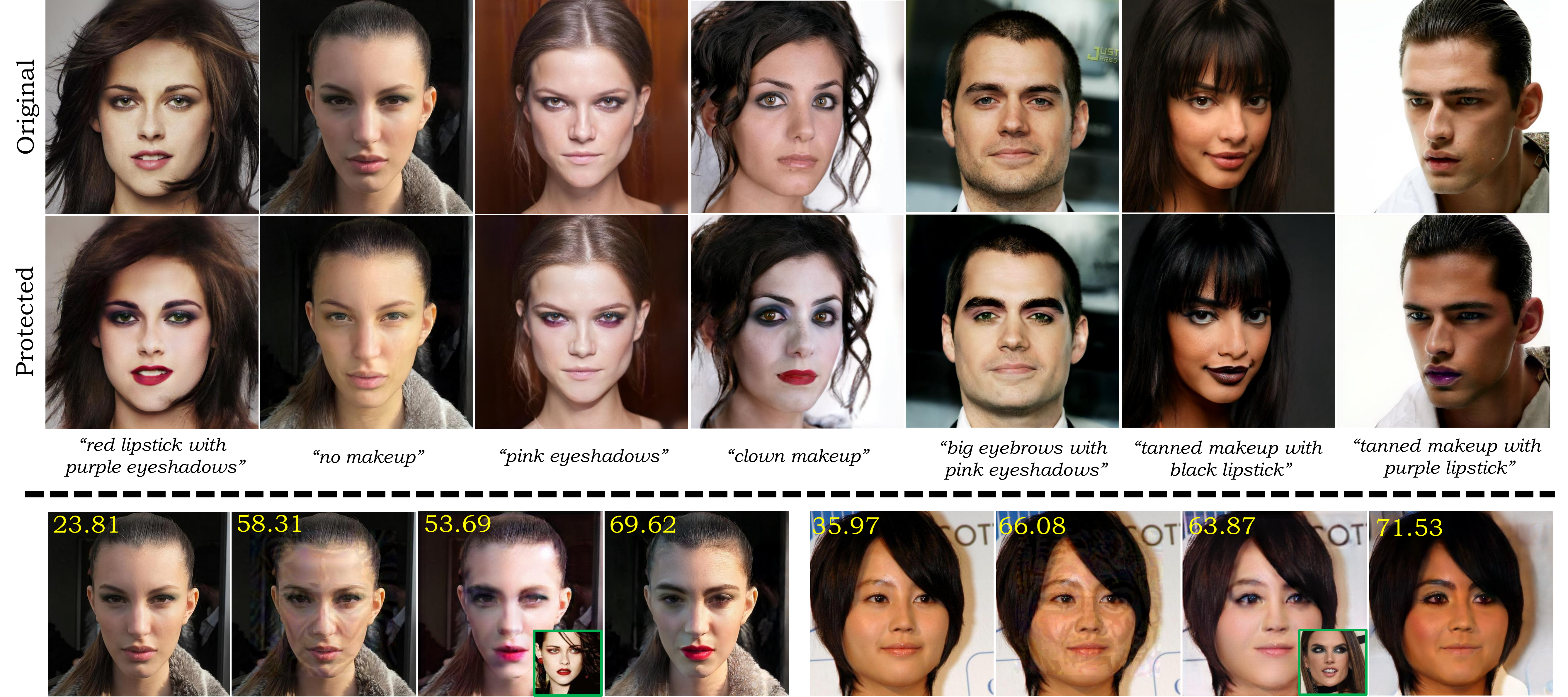}
 \put (6.5,-1.7) {\small Original}
 \put (16.3,-1.7) {\small TIP-IM~\cite{yang2021towards} }
 \put (27,-1.7) {\small AMT-GAN~\cite{hu2022protecting}}
  \put (41.4,-1.7) {\small Proposed}
  \put (54.8,-1.7) {\small Original}
 \put (65.9,-1.7) {\small TIP-IM~\cite{yang2021towards} }
 \put (76.3,-1.7) {\small AMT-GAN~\cite{hu2022protecting}}
  \put (90.8,-1.7) {\small Proposed}
\end{overpic}
\vspace{.1em}
        %\includegraphics[width=\linewidth]{figs/tit_fig.pdf}

    %\end{subfigure*}
    \begin{minipage}[t]{0.9\textwidth}
     \vspace{-1em}
       \captionsetup{type=figure}
  %\vspace{-0.8cm}
%   \captionof{figure}{\small Examples of text-guided adversaries to protect user facial privacy online.  Our approach crafts inconspicuous and transferable text-guided adversarial patterns to fool unauthorized face recognition systems without affecting user experience. Top row: original but vulnerable images; Bottom row: protected images. The user defined textual prompts
% to generate adversarial images are provided under each column. Yellow numbers are the confidence score (higher is better) returned by online commercial API, face++. For AMT-GAN reference makeup images are shown at the bottom corner. All images impersonate the target identity shown in the bottom right. Prompts for the last row are \textit{``pale makeup with red lipstick"} and \textit{``tanned makeup"} respectively.}
  
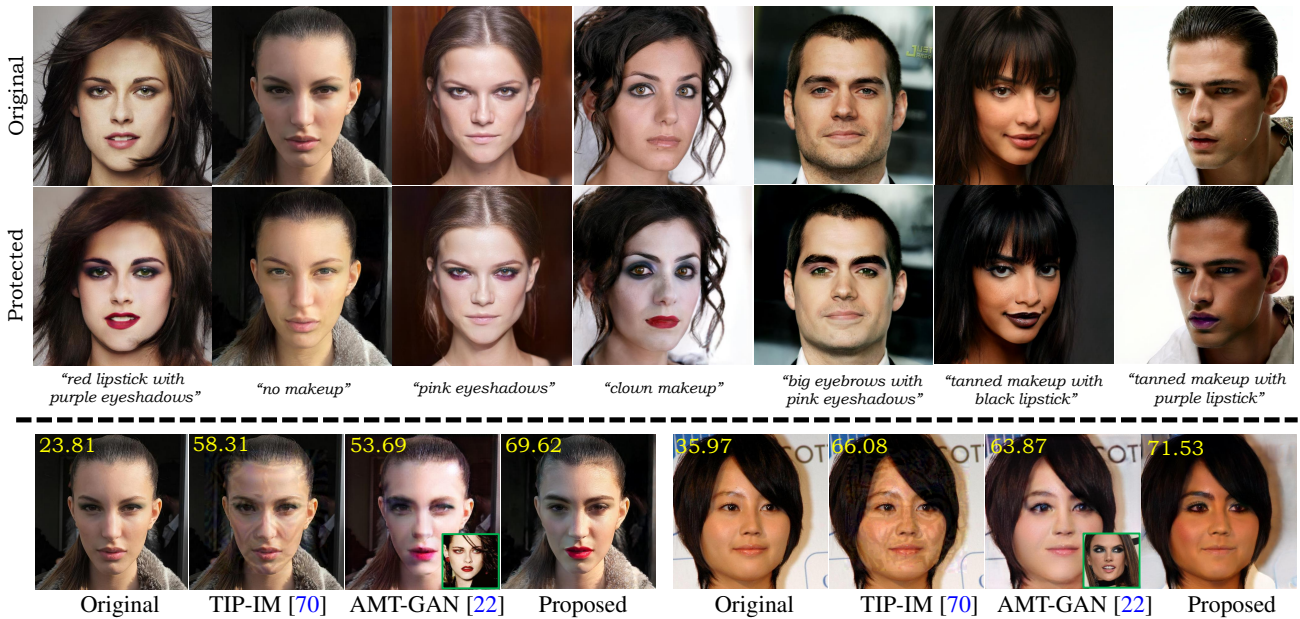
\captionof{figure}{\small The proposed approach crafts ``naturalistic'' and transferable text-guided adversarial faces to deceive black-box face recognition systems. First row shows original images that need to be protected and second row shows corresponding protected images along with the user-defined makeup text prompts that guide the adversarial search. Comparison against existing methods is shown in the third row. The yellow text represents the confidence score (higher is better) output by a commercial API (Face++), when matching the protected image against the target identity shown in the bottom right. The reference image used by \cite{hu2022protecting} for makeup transfer is shown at the bottom corner of the corresponding adversarial image.}
 %\vspace{0.5cm}
     %\caption{\small The results of the proposed approach....The results of the proposed approach....The results of the proposed approach....The results of the proposed approach....The results of the proposed approach....}
     \label{fig:title_fig}
    \end{minipage} 
    % \hspace{0.1em}
    \begin{minipage}[t]{0.085\textwidth}
    \strut\vspace*{-\baselineskip}\newline\includegraphics[width=\linewidth]{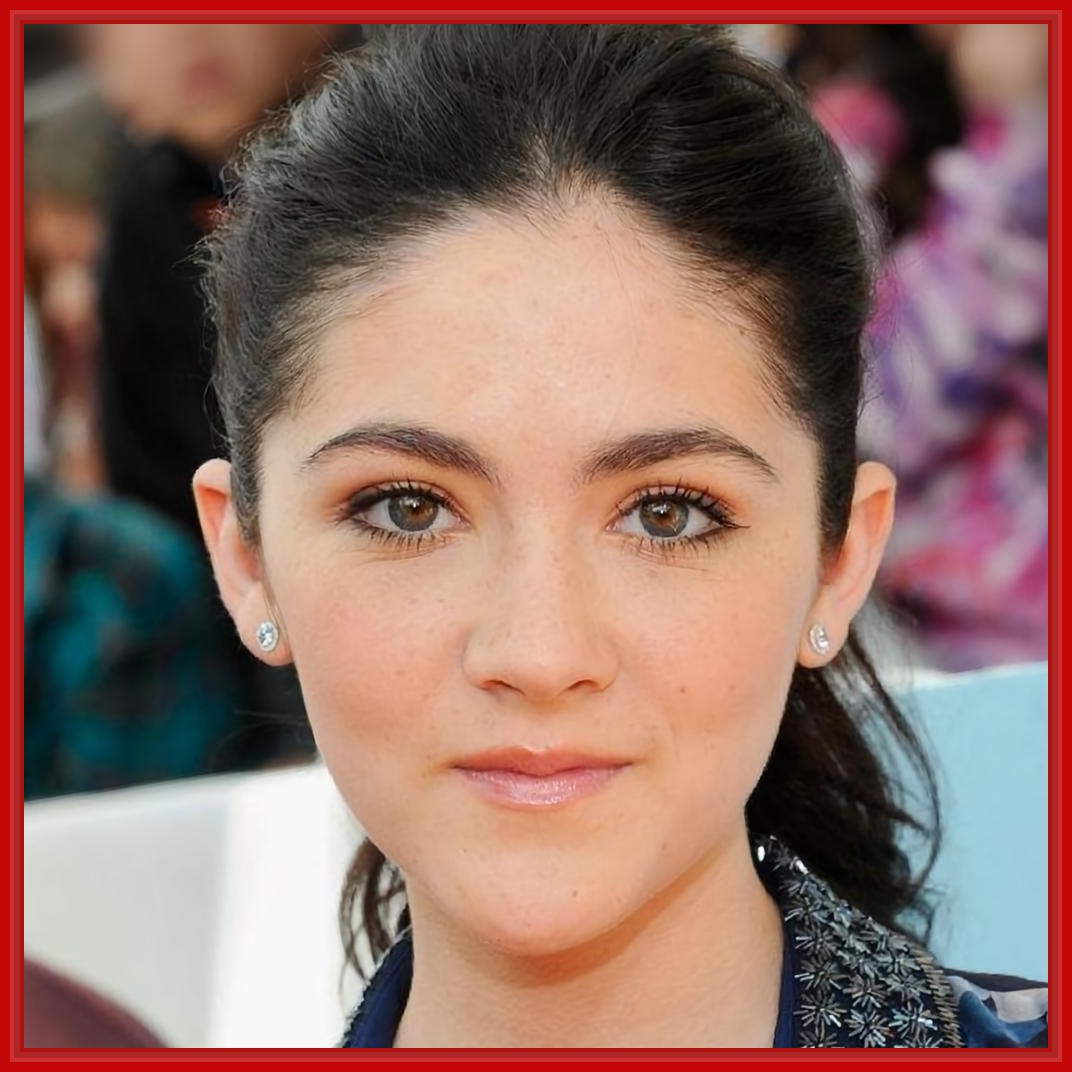}
      \put (-34,-7.5) {\small Target}
    \end{minipage} 
%\end{figure*}
 }]

%%%%%%%%% ABSTRACT
\begin{abstract}
The success of deep learning based face recognition systems has given rise to serious privacy concerns due to their ability to enable unauthorized tracking of users in the digital world. Existing methods for enhancing privacy fail to generate ``naturalistic'' images that can protect facial privacy without compromising user experience. We propose a novel two-step approach for facial privacy protection that relies on finding adversarial latent codes in the low-dimensional manifold of a pretrained generative model. The first step inverts the given face image into the latent space and finetunes the generative model to achieve an accurate reconstruction of the given image from its latent code. This step produces a good initialization, aiding the generation of high-quality faces that resemble the given identity. Subsequently, user-defined makeup text prompts and identity-preserving regularization are used to guide the search for adversarial codes in the latent space. Extensive experiments demonstrate that faces generated by our approach have stronger black-box transferability with an absolute gain of 12.06$\%$ over the state-of-the-art facial privacy protection approach under the face verification task. Finally, we demonstrate the effectiveness of the proposed approach for commercial face recognition systems. Our code is available at \href{https://github.com/fahadshamshad/Clip2Protect}{https://github.com/fahadshamshad/Clip2Protect}.
\end{abstract}
\vspace{-2em}

\section{Introduction} \label{sec:intro}

Deep learning based face recognition (\texttt{FR}) systems~\cite{parkhi2015deep,wang2021deep} have found widespread usage in multiple applications, including security~\cite{wang2017face}, biometrics~\cite{meden2021privacy}, and criminal investigation~\cite{phillips2018face}, outperforming humans in many scenarios~\cite{ranjan2018deep,wang2021deep,du2022elements}. Despite positive aspects of this technology, \texttt{FR} systems seriously threaten personal security and privacy in the digital world because of their potential to enable mass surveillance capabilities~\cite{ahern2007over,wenger2021sok}. For example, government and private entities can use \texttt{FR} systems to track user relationships and activities by scraping face images from social media profiles such as Twitter, Linkedin, and Facebook~\cite{hill2020secretive,heilweil2020world}. These entities generally use proprietary \texttt{FR} systems, whose specifications are unknown to the public (\emph{black box model}). Therefore, there is an urgent need for an effective approach that protects facial privacy against such unknown \texttt{FR} systems.
\iffalse
\begin{table}[t]
%\small
%\scriptsize 
%\normal
\caption{{\color{red} To be updated after experiments today}.}% observers. }
\footnotesize
\setlength{\tabcolsep}{5pt}
\centering
\begin{tabular}{l|ccccc}
\hline
& \tabincell{c}-& \tabincell{c}- & \tabincell{c}{Adv-Makeup}& \tabincell{c}{AMT-GAN} & Ours \\
\hline
%\tabincell{c}{Unknown gallery set} & No & No & No & No & No & \textbf{Yes} & No & No & \textbf{Yes}\\
\tabincell{c}{Target identity}  & \textbf{Yes} & \textbf{Yes} & \textbf{Yes} & \textbf{Yes} & \textbf{Yes}\\

\tabincell{c}{Black-box model} & No & Yes & \textbf{Yes} & \textbf{Yes} & \textbf{Yes}\\

\tabincell{c}{Natural outputs} & No & No & \textbf{Yes} & No & \textbf{Yes}\\

\tabincell{c}{Same faces} & \textbf{Yes} & \textbf{Yes} & \textbf{Yes} & \textbf{Yes} & \textbf{Yes}\\
\tabincell{c}{Unrestricted}  & No & No & \textbf{Yes} & \textbf{Yes} & \textbf{Yes}\\
\tabincell{c}{Text-prompt based}  & No & No & No & No & \textbf{Yes}\\
\hline
\end{tabular}
\vspace{-2ex}
%Note that \cite{dong2019efficient} considers the black-box face models with queries, but our method does not rely on such information.}
\label{tab:related-work}
\vspace{-3ex}
\end{table}
\fi
\begin{table}[t]
\begin{center}
\caption{\small Comparison among different facial privacy protection methods w.r.t. the natural outputs, black box setting, experiments under face verification and identification tasks, unrestricted (semantically meaningful), and more flexible text guided adversaries. }
\label{table:comparison}
\vspace{-2mm}
\setlength{\tabcolsep}{4.0pt}
\scalebox{0.76}{
\begin{tabular}{l | c   c   c  c }
\toprule[0.15em]
\rowcolor{mygray} &Adv-Makeup~\cite{DBLP:conf/ijcai/YinWYGKDLL21} &TIP-IM~\cite{yang2021towards}& AMT-GAN~\cite{hu2022protecting} & Ours \\ 
\midrule[0.15em]
Natural outputs  &  Yes & Partially  & Partially & Yes  \\
Black box  &  Yes & Yes  & Yes &Yes  \\
Verification  &  Yes & No  & Yes &Yes  \\
Identification  &  No & Yes  & No & Yes\\
Unrestricted  &  Yes & No  & Yes & Yes\\
Text guided  &  No & No  & No & Yes\\
\bottomrule[0.1em]
\end{tabular}}
\end{center}\vspace{-1.5em}
\end{table}

An ideal facial privacy protection algorithm must strike the right balance between naturalness and privacy protection~\cite{zhong2022opom,yang2021towards}. In this context, ``naturalness'' is defined as the \emph{absence of any noise artifacts that can be easily perceived by human observers} and the \emph{preservation of human-perceived identity}. ``Privacy protection'' refers to the fact that the protected image must be capable of \emph{deceiving a black-box malicious \texttt{FR} system}. In other words, the protected image must closely resemble the given face image and be artifact-free for a human observer, while at the same time fool an unknown automated \texttt{FR} system. Since failure to generate naturalistic faces can significantly affect user experience on social media platforms, it is a necessary pre-condition for adoption of a privacy-enhancement algorithm.

Recent works exploit adversarial attacks \cite{szegedy2013intriguing} %that can mislead \texttt{FR}
to conceal user identity by overlaying noise-constrained (bounded) adversarial perturbations on the original face image~\cite{zhang2020adversarial,cherepanova2020lowkey,shan2020fawkes}. 
%These perturbations are generally brought about by the adversarial optimization process in the image space, and may still sacrifice the visual quality for human perception due to the noise artifacts, thereby affecting user online experience~\cite{yang2021towards}.
%Although effective, 
Since the adversarial examples are generally optimized in the \textit{image space}, it is often difficult to simultaneously achieve naturalness and privacy~\cite{yang2021towards}. Unlike noise-based methods, unrestricted adversarial examples are not constrained by the magnitude of perturbation in the image space and have demonstrated better perceptual realism for human observers while being adversarially effective~\cite{xiao2018spatially,bhattad2019unrestricted,zhao2020towards,song2018constructing}. 

%Furthermore, these methods often rely on complete access to the target \texttt{FR} model (white-box setting), which is less practical in a real-world identity protection scenario.
%sacrifice visual quality for human perception due to gradient noise artifacts that can affect user's online experience. 

% In contrast to noise-based methods, unrestricted adversarial attacks (UAAs) are not constrained by the magnitude of the perturbation and have been shown to maintain perceptual realism for human observers~\cite{xiao2018spatially,bhattad2019unrestricted,zhao2020towards,song2018constructing}. 
% In contrast to gradient noise-based methods, generative approaches produce adversarial samples with perceptual realism for human observers~\cite{xiao2018spatially,bhattad2019unrestricted,zhao2020towards,song2018constructing}. 
%Recently, few efforts have been made to bring UAAs based generative approaches~\cite{poursaeed2018generative} to protect facial privacy. 
%Specifically,

Several efforts have been made to generate unrestricted adversarial examples that mislead \texttt{FR} systems (see Tab. \ref{table:comparison})~\cite{na2022unrestricted,kakizaki2019adversarial,yin2021adv,hu2022protecting}. Among these, adversarial makeup based methods~\cite{yin2021adv,hu2022protecting} are gaining increasing attention as they can embed adversarial modifications in a more natural way. These approaches use generative adversarial networks~\cite{DBLP:journals/corr/GoodfellowPMXWOCB14} (GANs) to 
adversarially transfer makeup from a given \emph{reference image} to the user's face image while impersonating a \emph{target identity}. However, existing techniques based on adversarial makeup transfer have the following limitations: (i) adversarial toxicity in these methods hamper the performance of the makeup transfer module, thereby resulting in unnatural faces with makeup artifacts (see Fig.~\ref{fig:title_fig}); (ii) the use of a reference image to define the desired makeup style affects the practicality of this approach; (iii) for every new target identity, these approaches require end-to-end retraining from scratch using large makeup datasets; and (iv) most of these methods primarily aim at impersonation of the target identity, whereas the desired privacy objective is dodging, \ie multiple images of the user's face scraped from different social media sites must not match with each other.

To mitigate the above problems, we propose a new approach to protect user facial privacy on online platforms (Sec. \ref{sec: Our Approach for Facial Privacy}). The proposed approach aims to search for \textbf{\textit{adversarial latent codes}} in a low-dimensional manifold learned by a generative model trained to generate face images ~\cite{karras2020analyzing,bermano2022state}. Our main contributions are:
\begin{itemize}
\itemsep0em
    \item \textbf{Facial Privacy-protection Framework Using Adversarial Latent Codes:} Given a face image, we propose a novel two-step method to search for adversarial latent codes, which can be used by a generative model (\eg, StyleGAN) to produce face images with high visual quality that matches human-perceived identity, while deceiving black-box \texttt{FR} systems.

    \item \textbf{Adversarial Makeup Transfer using Textual Prompts:} A critical component of the above framework is a technique for leveraging user-defined textual (makeup) prompts to traverse over the latent manifold of the generative model and find transferable adversarial latent codes. Our approach effectively hides attack information in the desired makeup style, without the need for any large makeup dataset or retraining of models for different target identities.

   \item \textbf{Identity Preserving Regularization:} We propose a regularizer that preserves identity-related attributes within the latent space of the generative model and ensures that the protected face image visually resembles the original face.
\end{itemize}

\noindent Extensive experiments (Sec. \ref{sec:results}) for both \emph{face verification} and  \emph{identification} scenarios demonstrate the effectiveness of our approach against black-box \texttt{FR} models and online commercial facial recognition APIs (Sec. \ref{sec:real-world}).  Furthermore, we provide detailed ablative analysis to dissect the performance of different components of our approach (Sec. \ref{sec:ablations}).

\section{Related Work} \label{sec:related_work}

% In this section, we provide related work regarding privacy protection of face images on social media.

%\textbf{Obfuscation-based methods.}

\noindent \textbf{Obfuscation Methods}: 
%The last decade has witnessed a rapid development of image-based facial privacy protection algorithms that can be broadly categorized into \textit{non-adversarial} and \textit{adversarial} methods. Among \textit{non-adversarial} approaches, 
Obfuscation is the most widely used technique~\cite{meden2021privacy} to protect user's facial privacy. Earlier obfuscation approaches typically degrade the quality of the original face image by applying simple operations such as masking~\cite{seneviratne2022does,wang2022privacy}, filtering~\cite{li2021deepblur,zhou2020personal}, and image transformations~\cite{wang2021gender,dabouei2019fast,liu2018face}. While these relatively simple obfuscation techniques are reasonable for surveillance applications, they are ill-suited for online/social media platforms where user experience is critical~\cite{oh2016faceless}. Though deep learning based obfuscation approaches generate more realistic images~\cite{sun2018natural,croft2022differentially,chen2021perceptual,tian2022fairness}, they  often result in a change of identity compared to the original image and occasionally produce undesirable artifacts~\cite{kuang2021effective,li2019anonymousnet,kuang2021unnoticeable}.

\noindent \textbf{Noise-based Adversarial Examples}: 
Adversarial attacks have been used to protect users from unauthorized \texttt{FR} models. Some methods ~\cite{cherepanova2020lowkey,shan2020fawkes} rely on data poisoning to deceive targeted \texttt{FR} models, but are less practical because access to the training data or the gallery set of the unknown \texttt{FR} system is often not available. % systems which is not practical in real world scenarios. %However, Fawkes assumes that practitioners train \texttt{FR} models on each individual’s data, that is not the case in practice where generally large pretrained Siamese networks are used. 
Other approaches have used game-theory perspective~\cite{oh2017adversarial} in white-box settings or person-specific privacy masks (one mask per person) to generate protected images at the cost of acquiring multiple images of the same user \cite{zhong2022opom}. In contrast, we aim to fool the \textit{black box} \texttt{FR} model using only single image.
%or end-cloud collaborated adversarial attack framework~\cite{zhang2020adversarial} where the original images are only available at the user end.
 In TIP-IM~\cite{yang2021towards}, targeted optimization was used to generate privacy masks against unknown \texttt{FR} models by introducing a naturalness constraint. While this approach provides effective privacy, it generates output images with perceptible noises that can affect the user experience~\cite{yang2021towards}. 
%In OPOM~\cite{zhong2022opom}, authors propose to generate person specific privacy masks to protect user in a more friendly way (one mask for every image of user), at the cost of acquiring multiple images of the same user. 

\noindent \textbf{Unrestricted Adversarial Examples}: Unrestricted adversarial attacks (UAAs) are not constrained by the perturbation norm and can induce large but semantically meaningful perturbations. These attacks have been extensively studied in image classification literature~\cite{xiao2018spatially,bhattad2019unrestricted,zhao2020towards,song2018constructing,liu2022towards,yuan2022natural} and it has been shown that outputs generated via UAAs are less perceptible to human observers as compared to noise-based adversarial attacks.
%%%%%%%%%%%%%%%%%%%%%%%%%%%%%%%%%%%%%%%%%%%%%%%%%%%%%%%%%%%%%%%%%%%%%%%%%%%%%%%%%%%%%%%%%
Motivated by this observation, patch-based unrestricted attacks have been proposed to generate wearable adversarial accessories like colorful glasses~\cite{sharif2019general}, hat~\cite{komkov2021advhat} or random patch~\cite{xiao2021improving} to fool the \texttt{FR} model, but such synthesized patches generally have weak transferability due to the limited editing region and the large visible pattern compromises naturalness and affects user experience. 
%%%%%%%%%%%%%%%%%%%%%%%%%%%%%%%%%%%%%%%%%%%%%%%%%%%%%%%%%%%%%%%%%%%%%%%%%%%%%%%%%%%%%%%%
Recently, generative models~\cite{salimans2016improved,isola2017image} have been leveraged to craft UAAs against \texttt{FR} models. However, these generative approaches are either designed for the white-box settings~\cite{zhu2019generating,poursaeed2021robustness} or show limited performance in query-free black-box settings~\cite{kakizaki2019adversarial}.
%%%%%%%%%%%%%%%%%%%%%%%%%%%%%%%%%%%%%%%%%%%%%%%%%%
Makeup-based UAAs~\cite{yin2021adv,guetta2021dodging} have also been proposed against \texttt{FR} systems by embedding the perturbations into a natural makeup effect. These makeup based attacks have also been exploited to protect the user privacy by applying adversarial makeup on the user face image~\cite{hu2022protecting}.  However, interference between adversarial perturbations and makeup transfer can produce undesirable makeup artifacts in the output images. Moreover, these attacks generally assume access to large makeup datasets for training models and require a reference makeup image.  In contrast, our approach finds adversarial faces on the natural image manifold in black-box setting \textit{via guidance from makeup text prompt}, which makes it less susceptible to artifacts (see Fig.~\ref{fig:title_fig}) and more practical. %, as will be shown in experiments. 
%{\color{red}may be one line about user friendliness as other approach require reference makeup image}.

\noindent \textbf{Vision-Language Modelling}: Cross-modal vision-language modelling has attracted significant attention in recent years~\cite{du2022survey}. OpenAI introduced CLIP~\cite{radford2021learning} that is trained on 400 million image-text pairs using contrastive objective and maps both image and text in a joint multi-modal embedding space. With powerful representation embedding of CLIP, several methods have been proposed to manipulate images with text-guidance. StyleCLIP~\cite{patashnik2021styleclip} and DiffusionCLIP~\cite{kim2022diffusionclip,nichol2021glide} leverage the powerful generative capabilities of StyleGAN and diffusion models to manipulate images with text prompts. Other similar works include HairCLIP~\cite{wei2022hairclip}, CLIP-NeRF~\cite{wang2022clip}, CLIPstyler~\cite{kwon2022clipstyler}, and CLIPDraw~\cite{frans2021clipdraw}.
%%%%%%%%%%%%%%%%%%%%%%%%%%%%%%%%%%%%%%%%%%%%%
While these methods focus on the text-guidance ability of CLIP, our approach aims to find the adversarial latent codes in a generative model's latent space for privacy protection against black-box \texttt{FR} models.%with high attack success rate. %Furthermore, we optimize only over the identity preserving latent codes {\color{red}refer experiment section?} , and thereby do not require facial recognition loss to preserve the identity of the output image as in the aforementioned works. {\color{red}may be one line about user friendliness}.

\section{Proposed Approach for Facial Privacy}
\label{sec: Our Approach for Facial Privacy}
\label{subsec: Preliminaries}
\begin{figure*}[h!]
\centering
\includegraphics[width=0.95\textwidth]{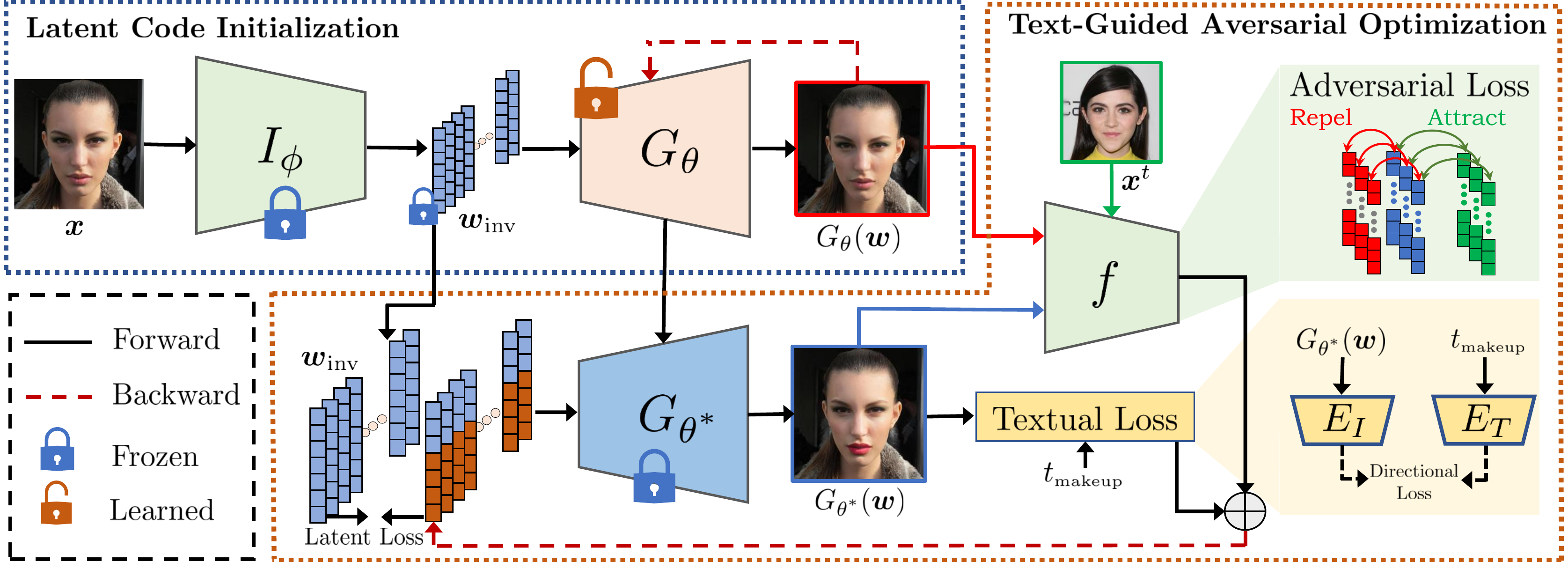}
\caption{Overall pipeline of the proposed approach to protect users facial privacy. Our proposed approach searches for the adversarial latent codes on the generative manifold to reconstruct an adversarial face that is  capable of fooling unknown FR systems for privacy protection. Our approach allows "makeup" editing in an adversarial manner through user defined textual prompts and thereby enhance the user's online experience. Our text-guided objective searches for such latent codes  while keeping the original identity preserved.}%  Our proposed approach consists of two stages: \textit{latent code initialization} and \textit{text-guided adversarial optimization}. Given the original image $\boldsymbol{x}$}
\label{fig:pipeline}
\end{figure*}

Our goal is to protect user facial privacy on online platforms against unknown (black-box) \texttt{FR} models without compromising on the user's online experience. The proposed approach finds protected faces by adversarially exploring the low-dimensional latent space of a pretrained generative model that is trained on natural face images. To avoid artifacts in the protected image, we restrict the search for adversarial faces close to the clean image manifold learned by the generative model. Moreover, we propose to optimize only over identity-preserving latent codes in the latent space. This effectively preserves human-perceived identity during attack while offering high privacy against automated systems. Further, we employ natural makeup-like perturbations via guidance from a text prompt, which provides more flexibility to the user compared to reference image-based adversarial makeup transfer \cite{hu2022protecting}.

%In the absence of such protection, attackers can utilize a black-box \texttt{FR} model to find the user identity and use it for malicious purposes.
%Existing approaches to protect facial privacy generally optimize the user face image in the \textit{image} space to craft adversarial examples that are misclassified by the malicious \texttt{FR} model. This image space optimization produces undesirable artifacts that can affect the user's online experience (see Fig. \ref{fig:title_fig}).
 % In comparison to existing makeup-based facial privacy protection methods, which use makeup images as a reference.  %The text-prompts not only provides users more flexibility but also make the proposed approach more resilient to the makeup artifacts caused by the semantic mismatch between reference makeup and original image due to adversarial toxicity.
%Lastly, although,  naively optimizing over the latent space produces natural face images, it can result in change in identity of the original image.

%Lastly large adversarial perturbation can 
%Moreover, to optimally balance the privacy-utility trade-off, we only optimize over the identity-preserving code in the latent space of the generative modelthese two competing objectives.

\subsection{Preliminaries}

Let $\boldsymbol{x} \in \mathcal{X} \subset \mathbb{R}^{n}$ denote the given original/real face image. Let $f(\boldsymbol{x}): \mathcal{X} \rightarrow \mathbb{R}^{d}$ be a \texttt{FR} model that extracts a fixed-length normalized feature representation. Let $\mathcal{D}(\boldsymbol{x}_1,\boldsymbol{x}_2) = D(f(\boldsymbol{x}_1),f(\boldsymbol{x}_2))$ be a distance metric that measures the dissimilarity between two face images $\boldsymbol{x}_1$ and $\boldsymbol{x}_2$ based on their respective representations $f(\boldsymbol{x}_1)$ and $f(\boldsymbol{x}_2)$.
%%%%%%%%%%%%%%%%%%%%%%%%%%%%%%%%%%%%%
Generally a \texttt{FR} system can operate in two modes: \emph{verification} and \emph{identification}. 
%%%%%%%%%%%%%%%%%%%%%%%%%%%%%%%%%%%%%
A face verification system predicts that two faces belong to the same identity if $\mathcal{D}(\boldsymbol{x}_1,\boldsymbol{x}_2) \leq \tau$, where $\tau$ is the system threshold. On the other hand, a (\emph{closed set}) face identification system compares the input image (probe) against a set of face images (gallery) and outputs the identity whose representation is most similar to that of the probe. Since the attacker can employ verification or identification to determine the user identity using black-box \texttt{FR} models, a protection approach should conceal the user's identity in both scenarios.

%Given a metric $\mathcal{D}(\boldsymbol{x}_1,\boldsymbol{x}_2) = \Vert f(\boldsymbol{x}_1) - f(\boldsymbol{x}_2) \Vert^2_2$, face verification aims to identify whether the two faces belong to the same identity based on a threshold on the distance metric. On the other hand, for face identification, the input image (probe) is compared against the set of face images (gallery) and recognized as the identity whose face feature representation is most similar to it. 
%%%%%%%%%%%%%%%%%%%%%%%%%%%%%%%%%%%%%

User privacy can be protected by misleading the malicious \texttt{FR} model through \emph{impersonation} or \emph{dodging} attacks. In the context of verification, impersonation (false match) implies that the protected face matches with the face of a specific target identity and dodging (false non-match) means that the protected face does not match with some other image of the same person. Similarly, for face identification, impersonation ensures that the protected image gets matched to a specified target identity in the gallery set, while dodging prevents the protected face from matching with images of the same person in the gallery.

%For face verification, impersonation aims to protect the original face image so that it is recognized as the ,
%of some other person 
%and dodging corresponds to generating a protected image that should not match with  
%Similarly, for face identification, impersonation ensures that the protected image matches the specified target identity
%of some other person 
%in the gallery set, and dodging aims to protect the image so that it should not match with the images of the same person in the gallery.

%%%%%%%%%%%%%%%%%%%%%%%%%%%%%%%%%%%
\noindent \textbf{Problem Statement}: Given the original face image $\boldsymbol{x}$, our goal is to generate a protected face image $\boldsymbol{x}^{p}$ such that $\mathcal{D}(\boldsymbol{x}^{p},\boldsymbol{x})$ is large (for successful dodging attack) and $\mathcal{D}(\boldsymbol{x}^{p},\boldsymbol{x}^{t})$ is small (for successfully impersonating a target face $\boldsymbol{x}^{t}$), where $\mathcal{O}(\boldsymbol{x}) \neq \mathcal{O}(\boldsymbol{x}^{t})$ and $\mathcal{O}$ is the oracle that gives the true identity labels. At the same time, we want to minimize $\mathcal{H}(\boldsymbol{x}^{p},\boldsymbol{x})$, where $\mathcal{H}$ quantifies the degree of unnaturalness introduced in the protected image $\boldsymbol{x}^{p}$ in relation to the original image $\boldsymbol{x}$. Formally, the optimization problem that we aim to solve is:
\begin{align} \label{eq:mai}
    \min_{\boldsymbol{x}^p} \mathcal{L}(\boldsymbol{x}^p) &= \mathcal{D}(\boldsymbol{x}^p,\boldsymbol{x}^t) - \mathcal{D}(\boldsymbol{x}^p,\boldsymbol{x})  \\
    \;& \text{s.t.} \; \mathcal{H}(\boldsymbol{x}^p,\boldsymbol{x}) \leq \epsilon \notag
    %\; &\text{s.t.} \; \Vert \boldsymbol{x} - \boldsymbol{x}^{p} \Vert_p \leq \epsilon \notag
\end{align}
\noindent where $\epsilon$ is a bound on the adversarial perturbation. For noise-based approach, $\mathcal{H}(\boldsymbol{x}^p,\boldsymbol{x}) = \Vert \boldsymbol{x} - \boldsymbol{x}^{p} \Vert_p$, where $\Vert\cdot\Vert_p$ denotes the $L_p$ norm. However, direct enforcement of the perturbation constraint leads to visible artifacts, which affects visual quality and user experience. Constraining the solution search space to a natural image manifold using an effective image prior can produce more realistic images. Note that the distance metric $\mathcal{D}$ is unknown since our goal is to deceive a black-box \texttt{FR} system.

\subsection{Makeup Text-Guided Adversarial Faces} \label{sec:textguided}
\label{subsec: Text-Guided Adversarial Faces}

\begin{figure}
     \centering
     \begin{subfigure}[b]{0.155\textwidth}
         \centering
         \includegraphics[width=\textwidth]{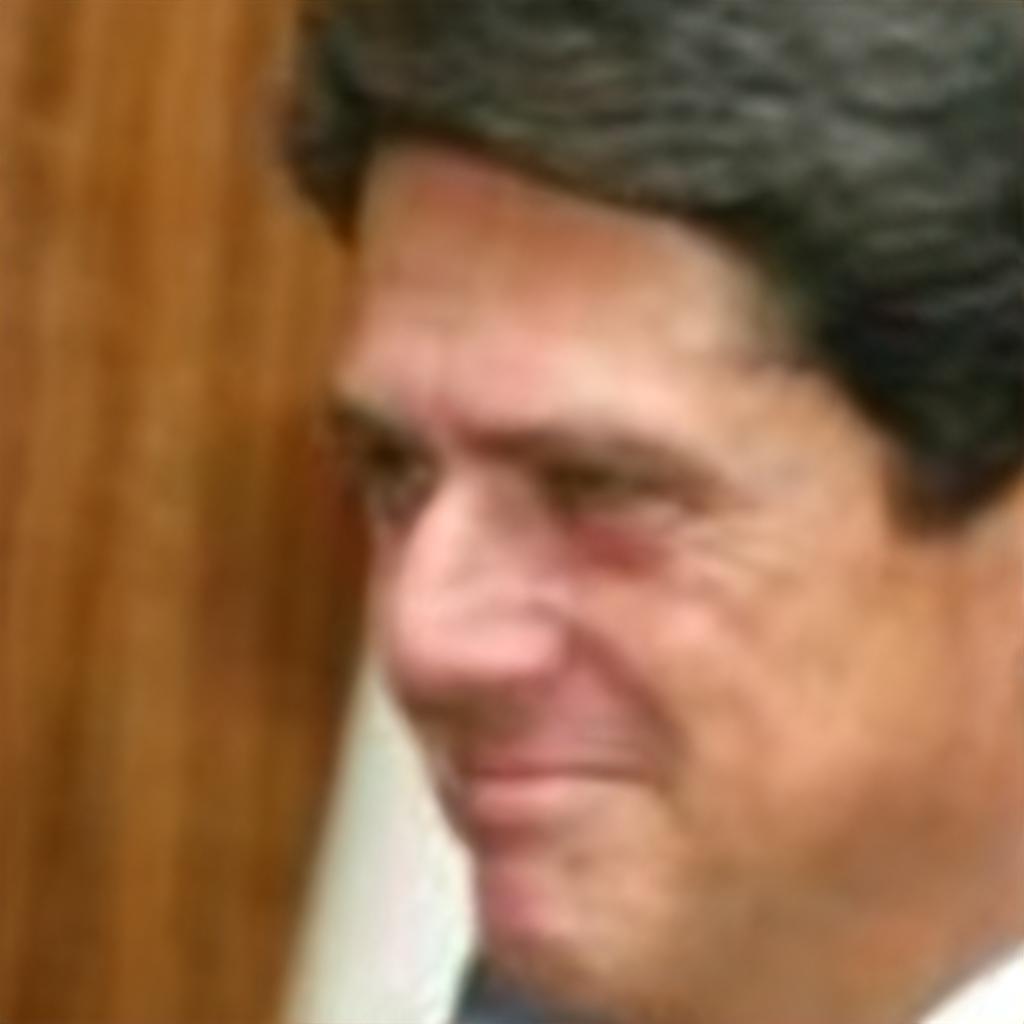}
         \caption{Original}
         \label{fig:y equals x}
     \end{subfigure}
     \hfill
     \begin{subfigure}[b]{0.155\textwidth}
         \centering
         \includegraphics[width=\textwidth]{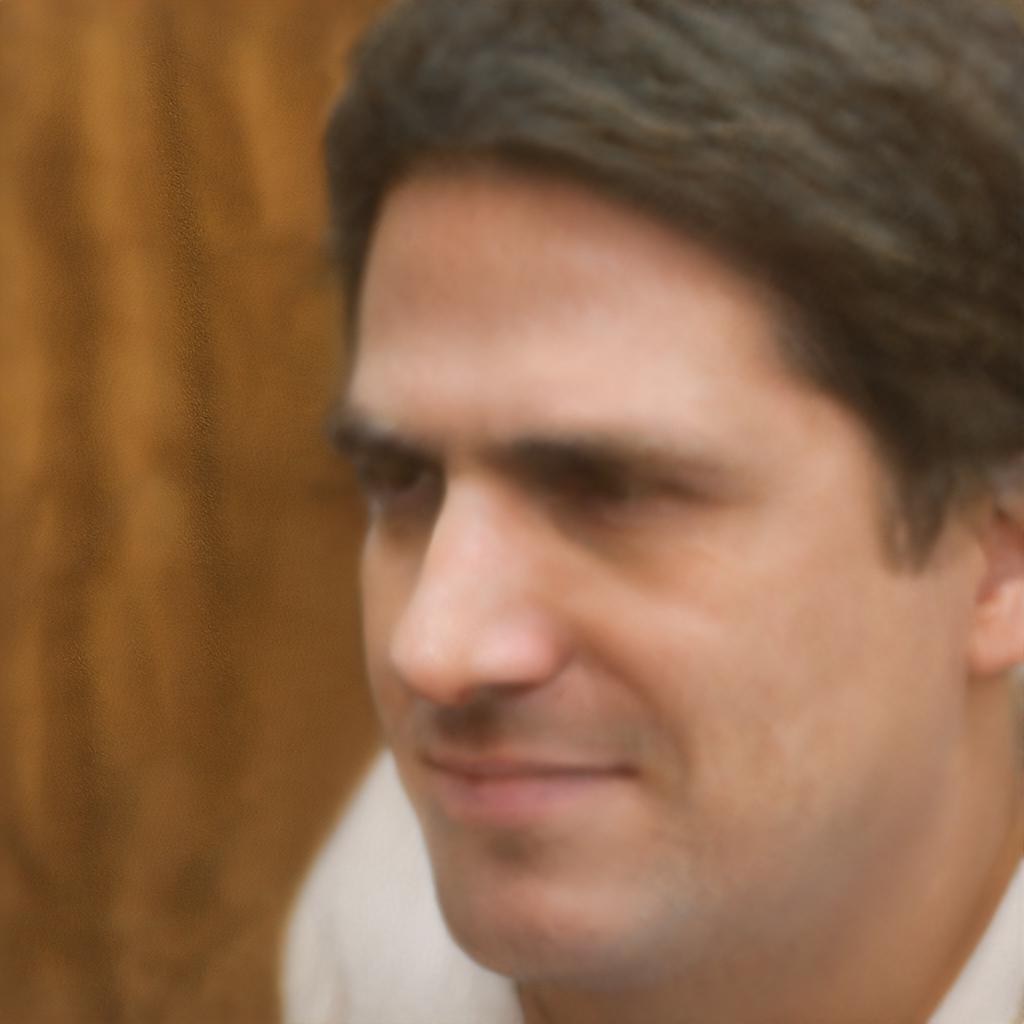}
         \caption{Encoder Inversion}
         \label{fig:three sin x}
     \end{subfigure}
     \hfill
     \begin{subfigure}[b]{0.155\textwidth}
         \centering
         \includegraphics[width=\textwidth]{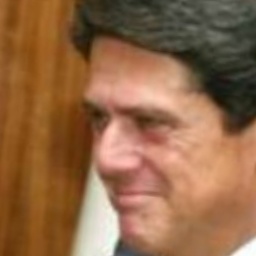}
         \caption{Generator finetuning}
         \label{fig:lfw_motiv}
     \end{subfigure}
     \vspace{-2em}
        \caption{Generator finetuning allows near-perfect reconstructions of LFW dataset sample. This is crucial for the online experience of users. Matching scores returned by Face++ API are 62.38 and 98.96 for encoder and generator-finetuned inversions, respectively.}
        \label{fig:g_finetune}
\end{figure}
%To address the aforementioned issues, we 
Our approach restricts the solution space of the protected face $\boldsymbol{x}^p$ to lie close to the clean face manifold $\mathcal{X}$. This manifold can be learned using a generative model trained on real human faces. Specifically, let $G_{\theta}(\boldsymbol{w}): \mathcal{W} \rightarrow \mathbb{R}^n$ denote the pretrained generative model with weights $\theta$, where $\mathcal{W}$ is the latent space. Our proposed approach consists of two stages: (i) latent code initialization (Sec. \ref{sec:gan_inversion}) and (ii) text-guided adversarial optimization (Sec. \ref{sec:adv_opt}). The overall pipeline of the proposed approach is shown in Fig. \ref{fig:pipeline}.

%,  \textit{\textbf{Latent Code Initialization}} (Sec. \ref{sec:gan_inversion}) and \textit{\textbf{text-guided adversarial optimization}} (Sec. \ref{sec:adv_opt}). 

%%%%%%%%%%%%%%%%%%%%%%%%%%%%%%%%%%%%%%%%%%%%%%%%%%%%%%%%%%%%%%%%%%

%%%%%%%%%%%%%%%%%%%%%%%%%%%%%%%%%%%%%%%%%%%%%%%%%%%%%%%%%%%%%%%%%
\subsubsection{Latent Code Initialization} \label{sec:gan_inversion}

The latent code initialization stage is based on GAN inversion, which aims to invert the original image $\boldsymbol{x}$ into the latent space $\mathcal{W}$, i.e., find a latent code $\boldsymbol{w}_{\text{inv}} \in \mathcal{W}$ such that 
$\boldsymbol{x}_{\text{inv}} = G_{\theta}(\boldsymbol{w}_{\text{inv}}) \approx \boldsymbol{x}$. To achieve this, we first use an encoder-based inversion called e4e~\cite{tov2021designing} to infer $\boldsymbol{w}_\text{inv}$ in $\mathcal{W}$ from $\boldsymbol{x}$ \ie $\boldsymbol{w}_\text{inv} = I_{\phi}(\boldsymbol{x})$, where $I_{\phi}:\mathcal{X}\rightarrow\mathcal{W}$ is the pretrained encoder with weights $\phi$ ({see Fig. \ref{fig:pipeline}}). 

We use StyleGAN trained on a high-resolution dataset of face images as the pretrained generative model $G_{\theta}$ due to its powerful synthesis ability and the disentangled structure of its latent space. A significant challenge during inversion is preserving the identity of the original image \ie $\mathcal{O}(\boldsymbol{x}) = \mathcal{O}(\boldsymbol{x}_{\text{inv}})$. Generally, optimization and encoder-based inversion approaches struggle to preserve identity after reconstruction~\cite{roich2022pivotal} (see Fig. \ref{fig:three sin x}). Moreover, when using these approaches, the inversion error can be large for out-of-domain face images with extreme poses and viewpoints, which are quite common in social media applications. Therefore, these approaches cannot be applied directly to invert $\boldsymbol{x}$. Instead, motivated by the recent observation \cite{roich2022pivotal} that slight changes to the pretrained generator weights do not harm its editing abilities while achieving near-perfect reconstructions, we finetune the pretrained generator weights $\theta$ instead of the encoder weights $\phi$. Specifically, we fix $\boldsymbol{w}_\text{inv}= I_{\phi}(\boldsymbol{x})$ and fine-tune $G_{\theta}$ using the following loss: 
\begin{equation} \label{eq:ganinv}
\small
    \theta^{*} =  \underset{\theta}{\arg\min}\; \notag \mathcal{L}_{\text{LPIPS}}(\boldsymbol{x},G_{\theta}(\boldsymbol{w}_{\text{inv}})) + \lambda_2\mathcal{L}_2(\boldsymbol{x},G_{\theta}(\boldsymbol{w}_{\text{inv}})), 
\end{equation}
where $\mathcal{L}_{\text{LPIPS}}$ is the perceptual loss and $\mathcal{L}_2$ denotes the pixel-wise similarity. The final inverted image $\boldsymbol{x}^*_\text{inv}$ (see Fig. \ref{fig:lfw_motiv}) can be obtained by performing a forward pass of $\boldsymbol{w}_{\text{inv}}$ through fine-tuned generator \ie $\boldsymbol{x}^*_\text{inv} = G_{\theta^{*}}(\boldsymbol{w}_{\text{inv}})$.

\subsubsection{Text-guided adversarial optimization} \label{sec:adv_opt}

%The careful design of the loss functions is crucial to the success of the proposed approach. 
Given the inverted latent code $\boldsymbol{w}_{\text{inv}}$ and fine-tuned generator $G_{\theta^{*}}(.)$, our goal is to adversarially perturb this latent code $\boldsymbol{w}_{\text{inv}}$ in the low-dimensional generative manifold $\mathcal{W}$ to generate a protected face that fools the black-box \texttt{FR} model, while imitating the makeup style of the text prompt $t_{\text{makeup}}$. 

To achieve these objectives, we investigate the following questions: (i) how to effectively extract makeup style information from $t_\text{makeup}$ and apply it to the face image $\boldsymbol{x}$ in an \textit{adversarial} manner?, (ii) how to regularize the optimization process so that the output face image is not qualitatively impaired?, (iii) how to craft effective adversarial perturbations that mislead \textit{black-box} \texttt{FR} models?, and (iv) how to preserve the human-perceived identity $\mathcal{O}(\boldsymbol{x})$ of the original face image while ensuring high privacy?

The first issue can be addressed by aligning the output adversarial image with the text prompt $t_{\text{makeup}}$ in the embedding space of a pretrained vision-language model. The second issue is addressed by enforcing the adversarial latent code to remain close to initialization $\boldsymbol{w}_{\text{inv}}$. The third issue is solved by crafting transferable text-guided adversarial faces on a white-box surrogate model (or an ensemble of models) with the goal of boosting the fooling rate on the black-box \texttt{FR} model. Finally, we leverage the disentangled nature of latent space in the generative model and incorporate an identity-preserving regularization to effectively maintain the original visual identity. We now present the details of the loss functions used to incorporate the above ideas.

\noindent \textbf{Textual Loss}: A key ingredient of the proposed approach is text-based guidance to inconspicuously hide the adversarial perturbations into the makeup effect. This can be naively achieved by aligning the representation of $t_{\text{makeup}}$ and the adversarial face $G_{\theta^*}(\boldsymbol{w})$ in the common embedding space of a pre-trained vision-language model (\eg CLIP~\cite{radford2021learning}). However, this approach will transform the \textit{whole} output image to follow the makeup style of $t_{\text{makeup}}$, which results in low diversity. Therefore, we use a directional CLIP loss that aligns the CLIP-space direction between the text-image pairs of the original and adversarial images. Specifically,
\vspace{-0.8em}
\begin{align} \label{eq:dir}
\mathcal{L}_{\text{clip}} &=   1-\frac{\Delta I\cdot \Delta T}{|\Delta I||\Delta T|},
\end{align}
%Specifically, we use the the CLIP model~\cite{radford2021learning}, which consists of a text encoder $E_T$ and an image encoder $E_I$ to map their respective inputs into a shared encoding space, where similarity is measured by cosine distance. Thus, the loss function is:
%\begin{equation} \label{eq:global}
%     \mathcal{L}_{global} = 1 - \text{cos}[E_{I}(G_{\theta^*}(\boldsymbol{w})),E_{T}(t_{\text{makeup}})].
%\end{equation}
%\noindent The above loss function transforms the \textit{whole} output image to follow the makeup style of $t_{\text{makeup}}$, which results in low diversity. 
\noindent where $\Delta T =  E_{T}(t_{\text{makeup}}) - E_{T}(t_{\text{src}})$ and $\Delta I =  E_{I}(G_{\theta^*}(\boldsymbol{w})) - E_{I}(\boldsymbol{x})$. Here, $E_T$ and $E_I$ are the text and image encoders of the CLIP model and $t_{\text{src}}$ is the semantic text of the input image $\boldsymbol{x}$. Since we are dealing with faces, $t_{\text{src}}$ can be simply set as ``face''. This loss localizes makeup transfer (\eg red lipstick) without affecting privacy.

%\noindent 
\noindent \textbf{Adversarial Loss}: Our goal is to traverse over the latent space $\mathcal{W}$ to find adversarial latent codes on the generative manifold whose face feature representation lies close to that of target image and far away from the original image itself  \ie $\mathcal{D}(\boldsymbol{x}^{p},\boldsymbol{x}) > \mathcal{D}(\boldsymbol{x}^{p},\boldsymbol{x}^{t})$. Hence, the adversarial loss is:
%The adversarial loss term aims to generate adversarial output examples which can be recognized as the specified target identity (i.e. impersonation attack). Specifically, the loss minimize the distance between the embeddings of the adversarial output image $\mathcal{G}(\boldsymbol{x})$ and the target image $\boldsymbol{x}_t$. These embeddings are extracted via a pre-trained face feature extractor denoted as $\mathcal{F}$. 
%Integrating the generative prior constraint in Eq. \ref{eq:mai} and initializing the  latent code to be optimized with $w_{\text{inv}}$ we get:
\begin{equation} \label{eq:reform}
%\small
    \mathcal{L_{\text{adv}}} = \mathcal{D}(G_{\theta^*}(\boldsymbol{w}),\boldsymbol{x}^t) - \mathcal{D}(G_{\theta^*}(\boldsymbol{w}),\boldsymbol{x}), 
\end{equation}

\noindent where $\mathcal{D}(\boldsymbol{x}_1,\boldsymbol{x}_2) = 1-\text{cos}[f(\boldsymbol{x}_1),f(\boldsymbol{x}_2))]$ is the cosine distance. Since the malicious \texttt{FR} model is unknown in the \textit{black-box} setting, Eq. \ref{eq:reform} cannot be solved directly. Instead, following AMT-GAN~\cite{hu2022protecting}, we perform adversarial optimization on an ensemble of white-box surrogate models to imitate the decision boundary of the unknown \texttt{FR} model.
%%%%%%%%%%%%%%%%%%%%%%%%%%%%%%%%%%%%%%%%%%%%%%%%%%%%%%%%%%%

\noindent \textbf{Identity Preservation Loss}: The optimization over the generative manifold ensures that the protected image $\boldsymbol{x}^p$ is natural \ie artifact-free, however, it does not explicitly enforce the protected image to preserve the identity of the original image with respect to the human observer. To mitigate the issue, we take advantage of the semantic control exhibited by StyleGAN in its latent space. The latent code $\boldsymbol{w} \in \mathcal{W}$ impacts image generation by controlling different level of semantics in the output image. Specifically, latent codes corresponding to the initial layers of StyleGAN control high-level aspects such as pose, general hairstyle, and face shape \cite{karras2020analyzing}. Adversarially perturbing these latent layers can change these attributes, resulting in a change of identity (see Sec. \ref{sec:ablations}). Latent codes corresponding to deeper layers of StyleGAN are associated with fine-level control such as makeup style~\cite{bermano2022state}. %while keeping other attributes intact.
%%%%%%%%%%%%%%%%%%%%%%%%%%%%%%%%%%%%%%%%%%%
Therefore, we perturb only those latent codes associated with deeper layers of StyleGAN, thereby restricting the adversarial faces to the identity preserving manifold. We further constrain the latent code to stay close to its initial value $\boldsymbol{w}_{\text{inv}}$ using the following regularization:
\begin{equation}
    \mathcal{L}_{\text{latent}} = \Vert (\boldsymbol{w} \odot \boldsymbol{m}_{id}) - (\boldsymbol{w}_{\text{inv}} \odot \boldsymbol{m}_{id})\Vert_2, 
\end{equation}

\noindent where $\odot$ denotes element-wise product and $\boldsymbol{m}_{id}$ is an identity preservation mask that is $0$ for the initial layers and $1$ only for the deeper layers of the latent code. StyleGAN has 18 layers, each having a dimension of 512. The identity preservation mask is set to 1 only from layer 8 to 18.
%%%%%%%%%%%%%%%%%%%%%%%%%%%%%%%%%%%%%%%%
Finally, combining the three loss functions, we have
\begin{equation}
    \mathcal{L}_{\text{total}} =  \lambda_{\text{adv}} \mathcal{L}_{\text{adv}} +  \lambda_{\text{clip}}  \mathcal{L}_{\text{clip}} + \lambda_{\text{latent}} \mathcal{L}_{\text{latent}},
\end{equation}
where $\lambda_{\text{adv.}}$, $\lambda_{\text{clip}}$, and $\lambda_{\text{latent}}$ are hyperparameters. Note that $\mathcal{L}_{\text{adv}}$ accounts for the adversarial objective in Eq. \ref{eq:mai}, while the text-guided makeup transfer ($\mathcal{L}_{\text{clip}}$) and identity-preserving regularization ($\mathcal{L}_{\text{latent}}$) implicitly enforce the naturalness constraint in Eq. \ref{eq:mai}.

%by utilizing powerful synthesis ability of the pretrained StyleGAN to craft image-specific make-up adversarial attack guided by a text prompt. 

%\clearpage
\section{Experiments} \label{sec:experiments}

\noindent \textbf{Implementation details:} In all experiments, we use StyleGAN2 pretrained on the FFHQ face dataset as our generative model. For adversarial text guidance, we use a vision transformer-based CLIP model.
%%%%%%%%%%%%%%%%%%%%%%%%%%%%%%%%%%%%%%%%%%%
For generator fine-tuning in the latent code initialization step, we use 450 iterations with value of $\lambda_2$ in Eq. \ref{eq:ganinv} set to 0.5.
%%%%%%%%%%%%%%%%%%%%%%%%%%%%%%%%%%%%%%%%%%%
For the makeup text input, we collect 40 text prompts based on the makeup style of diverse nature (details in supplementary material). 
%%%%%%%%%%%%%%%%%%%%%%%%%%%%%%%%%%%%%%%%%%%%%%
For adversarial optimization, we use an Adam optimizer with $\beta_1$ and $\beta_2$ set to 0.9 and 0.999, respectively, and a learning rate of 0.01. We run the optimizer for 50 iterations to craft protected faces. %During optimization, 
We set the value of $\lambda_{\text{adv}}$, $\lambda_{\text{clip}}$, and $\lambda_{\text{latent}}$ to 1, 0.5, and 0.01, respectively.
%%%%%%%%%%%%%%%%%%%%%%%%%%%%%%%%%%%%%%%%%%%%%%
All our experiments are conducted on a A100 GPU with 40 GB memory.

\textbf{Datasets:} We perform experiments for both face verification and identification settings. \underline{\textit{Face verification}}: We use CelebA-HQ~\cite{karras2018progressive} and LADN~\cite{gu2019ladn} for the impersonation attack. We select subset of 1,000 images from CelebA-HQ and report average results over 4 target identities provided by~\cite{hu2022protecting}. Similarly, for LADN, we divide the 332 images available into 4 groups, where images in each group aim to impersonate the target identities provided by \cite{hu2022protecting}. For dodging attack, we use CelebA-HQ~\cite{karras2018progressive} and LFW~\cite{huang2008labeled} datasets.  Specifically, we select 500 subjects at random and each subject has a pair of faces.
%%%%%%%%%%%%%%%%%%%%%%%%%%%%%%%%%%%%%%%%%%%%%%%%%%%%%%%%%%%%%%
\textit{\underline{Face identification}}: For impersonation and dodging, we use CelebA-HQ~\cite{karras2018progressive} and LFW~\cite{huang2008labeled} as our evaluation set. For both datasets, we randomly select 500 subjects, each with a pair of faces. We assign one image in the pair to the gallery set and the other to the probe set. Both impersonation and dodging attacks are performed on the probe set. For impersonation, we insert 4 target identities provided by \cite{hu2022protecting} into the gallery set. A more detailed description of all datasets and pre-processing steps is provided in the supplementary material.
%%%%%%%%%%%%%%%%%%%%%%%%%%%%%%%%%%%%%%%%%%%%%%%%%%%%%%%%%%%%%%
\begin{table*}
\begin{center}
\caption{\small Protection success rate (PSR \%) of \textit{black-box} impersonation attack under the face verification task. For each column, the other three \texttt{FR} systems are used as surrogates to generate the protected faces. }
\label{table:verification_impersonation}
\vspace{-2mm}
\setlength{\tabcolsep}{7.0pt}
\scalebox{0.78}{
\begin{tabular}{l || c c c c || c c c c || c }
\toprule[0.15em]
\rowcolor{mygray} \textbf{Method} & \multicolumn{4}{c||}{\textbf{CelebA-HQ}}&\multicolumn{4}{c||}{\textbf{LADN-Dataset}}&\multicolumn{1}{c}{\textbf{Average}} \\
\rowcolor{mygray}  & IRSE50 & IR152 & FaceNet& MobileFace & IRSE50 & IR152 & FaceNet& MobileFace &  \\
\midrule[0.15em]
Clean & 7.29  & 3.80  & 1.08  & 12.68  & 2.71  & 3.61  & 0.60  & 5.11  & 4.61   \\
Inverted & 5.57 & 2.77 & 0.60 & 13.32 & 6.80 & 4.51 & 0.25 & 11.66 & 5.68  \\
PGD~\cite{DBLP:conf/iclr/MadryMSTV18}&  36.87  & 20.68 & 1.85    & 43.99      & 40.09   & 19.59  & 3.82     & 41.09 & 25.60 \\
MI-FGSM~\cite{DBLP:conf/cvpr/DongLPS0HL18} & 45.79  & 25.03 & 2.58     & 45.85      & 48.90    & 25.57  & 6.31     & 45.01 & 30.63 \\
TI-DIM~\cite{DBLP:conf/cvpr/DongPSZ19} & 63.63&36.17& 15.30&57.12& 56.36&34.18& 22.11&48.30& 41.64 \\
$\text{Adv-Makeup}_{\text{(IJCAI'21)}}$~\cite{DBLP:conf/ijcai/YinWYGKDLL21} & 21.95  & 9.48  & 1.37    & 22.00      & 29.64   & 10.03  & 0.97     & 22.38& 14.72  \\
$\text{TIP-IM}_{\text{(ICCV'21)}}$~\cite{yang2021towards}  & 54.40 & 37.23 & 40.74& 48.72 & 65.89 & 43.57 & \textbf{63.50} & 46.48 & 50.06 \\
$\text{AMT-GAN}_{\text{(CVPR'22)}}$~\cite{hu2022protecting}  & 76.96 & 35.13 & 16.62 & 50.71 & 89.64 & 49.12 & {32.13} & {72.43} & 52.84  \\
\midrule
\rowcolor{orange!6} Ours & \textbf{81.10} &\textbf{48.42} & \textbf{41.72} & \textbf{75.26} & \textbf{91.57} & \textbf{53.31} & 47.91 & \textbf{79.94} & \textbf{64.90}  \\
\bottomrule[0.1em]
\end{tabular}}
\end{center}\vspace{-1.5em}
\end{table*}

\begin{table*}
\begin{center}
\caption{\small Protection success rate (PSR \%) of \textit{black-box} dodging (top) and impersonation (bottom) attacks under the face identification task for LFW dataset~\cite{huang2008labeled}. For each column, the other three \texttt{FR} systems are used as surrogates to generate the protected faces. R1-U: Rank-1-Untargeted, R5-U: Rank-5-Untargeted,  R1-T: Rank-1-Targeted, R5-T: Rank-5-Targeted. }
\label{table:identification}
\vspace{-2mm}
\setlength{\tabcolsep}{13.0pt}
\scalebox{0.75}{
\begin{tabular}{l || c c || c c || c c || c c || c c }
\toprule[0.15em]
\rowcolor{mygray} \textbf{Method} & \multicolumn{2}{c||}{\textbf{IRSE50}}&\multicolumn{2}{c||}{\textbf{IR152}}&\multicolumn{2}{c||}{\textbf{FaceNet}}&\multicolumn{2}{c||}{\textbf{MobileFace}}&\multicolumn{2}{c}{\textbf{Average}} \\
\rowcolor{mygray}  & R1-U & R5-U & R1-U & R5-U & R1-U & R5-U & R1-U & R5-U & R1-U & R5-U  \\
\midrule[0.15em]
MI-FGSM~\cite{DBLP:conf/cvpr/DongLPS0HL18}  & 70.2 & 42.6 & 58.4 & 41.8 & 59.2 & 34.0 & 68.0 & 47.2 & 63.9 &41.4  \\
TI-DIM~\cite{DBLP:conf/cvpr/DongPSZ19} & 79.0 &51.2 & 67.4 & 54.0 & 74.4 & 52.0 & 79.2 & 61.6 & 75.0&54.7 \\
$\text{TIP-IM}_{\text{(ICCV'21)}}$~\cite{yang2021towards} & 81.4 & 52.2& 71.8 & 54.6 & 76.0 & 49.8 & 82.2 & 63.0 & 77.8&54.9 \\
\midrule
\rowcolor{orange!6} Ours & \textbf{86.6} &\textbf{59.4} & \textbf{73.4} & \textbf{56.6} & \textbf{83.8} & \textbf{51.2} & \textbf{85.0} & \textbf{66.8} & \textbf{82.2} & \textbf{58.5}  \\
%\bottomrule[0.1em]
\midrule[0.15em]
%%%%%%%%%%%%%%%%%%%%%%%%%%%%%%%%%%%%%%%%%%%%%%%%%%%%%%%%%%%%%%%%%%%%%%%%%%%%%%%%%%%%%%%%%%%%%%%%%%%%%%%%%%%%%%%%%%%%%%%%%%%%%%%%%%%
\rowcolor{mygray}  & R1-T & R5-T & R1-T & R5-T & R1-T & R5-T & R1-T & R5-T & R1-T & R5-T  \\
\midrule[0.15em]
MI-FGSM~\cite{DBLP:conf/cvpr/DongLPS0HL18}  & 4.0 & 10.2 & 3.2 & 14.2 & 9.0 & 18.8 & 8.4 & 22.4& 6.15 & 16.4\\
TI-DIM~\cite{DBLP:conf/cvpr/DongPSZ19} & 4.0 &13.6 & 7.8& 19.6 & 18.0 & 32.8 & 21.6 & 39.0 & 12.85& 26.25 \\
$\text{TIP-IM}_{\text{(ICCV'21)}}$~\cite{yang2021towards} & 8.0 &28.2& 11.6 & 31.2 & 25.2 & \textbf{56.8} & 34.0 & 51.4 & 19.7& 41.9\\
\midrule
\rowcolor{orange!6} Ours & \textbf{11.2} &\textbf{37.8} &\textbf{16.0} & \textbf{51.2} & \textbf{27.4} & 54.0 & \textbf{39.0} & \textbf{61.2} &\textbf{23.4} &\textbf{51.05} \\
\bottomrule[0.1em]
\end{tabular}}
\end{center}\vspace{-1.5em}
\end{table*}

\textbf{Target Models:} We aim to protect user facial privacy by attacking four \texttt{FR} model with diverse back bones in the black-box settings. The target models include IRSE50~\cite{hu2018squeeze}, IR152~\cite{deng2019arcface}, FaceNet~\cite{schroff2015facenet}, and MobileFace~\cite{chen2018mobilefacenets}. Following standard protocol, we align and crop the face images using MTCNN~\cite{zhang2016joint} before giving them as input to \texttt{FR} models.
Further, we also report privacy protection performance based on commercial \texttt{FR} API including Face++ and Tencent Yunshentu \texttt{FR} platforms.

\textbf{Evaluation metrics:} Following \cite{yang2021towards}, we use protection success rate (PSR) to evaluate the proposed approach. PSR is defined as the fraction of protected faces missclassified by the malicious \texttt{FR} system. To evaluate PSR, we use the thresholding and closed set strategies for face verification and identification, respectively.
For face identification, we also use Rank-N targeted identity success rate (\textit{Rank-N-T}) and untargeted identity success rate (\textit{Rank-N-U}), where \textit{Rank-N-T} means that target image $\boldsymbol{x}^t$ will appear at least once in the top N candidates shortlisted from the gallery and \textit{Rank-N-U} implies that the top $N$ candidate list does not have the same identity as that of original image $\boldsymbol{x}$. 
We also report results of PSNR (dB), SSIM, and FID~\cite{heusel2017gans} scores to evaluate the imperceptibility of method. Large PSNR and SSIM~\cite{wang2004image} indicates better match with the original images, while low FID score indicates more realistic images. For commercial APIs, we directly report the confidence score returned by the respective servers.

\textbf{Baseline methods:} We compare our approach with recent noise-based and makeup based facial privacy protection approaches. Noise based methods include PGD~\cite{DBLP:conf/iclr/MadryMSTV18}, MI-FGSM~\cite{DBLP:conf/cvpr/DongLPS0HL18}, TI-DIM~\cite{DBLP:conf/cvpr/DongPSZ19}, and TIP-IM~\cite{yang2021towards}, whereas makeup-based approaches are Adv-Makeup~\cite{DBLP:conf/ijcai/YinWYGKDLL21} and AMT-GAN~\cite{hu2022protecting}. We want to highlight that TIP-IM and AMT-GAN are considered the state-of-the-art (SOTA) for face privacy protection against \textit{black-box} \texttt{FR} systems in noise-based and unrestricted settings, respectively. TIP-IM also incorporate multi-target objective in its optimization to find the optimal target image among multiple targets. For fair comparison, we use its single target variant.
%%%%%%%%%%%%%%%%%%%%%%%%%%%%%%%%%%%%%%%%%%%%%%%%%%%%%%
%%%%%%%%%%%%%%%%%%%%%%%%%%%%%%%%%%%%%%%%%%%%%%%%%%%%%5
%%%%%%%%%%%%%%%%%%%%%%%%SSIM Results%%%%%%%%%%%%%%%%%%
%%%%%%%%%%%%%%%%%%%%%%%%%%%%%%%%%%%%%%%%%%%%%%%%%%%%%%%%%%%%%%
\begin{SCtable}[\sidecaptionrelwidth][t]
\label{table:fid}
\vspace{-2mm}
\setlength{\tabcolsep}{7.0pt}
\scalebox{0.75}{
\begin{tabular}{l | c  |c }
\toprule[0.15em]
\rowcolor{mygray} \textbf{Method} & FID $\downarrow$  & PSR Gain $\uparrow$ \\
\midrule[0.15em]
Adv-Makeup~\cite{DBLP:conf/ijcai/YinWYGKDLL21} & 4.23  &  0    \\
TIP-IM~\cite{yang2021towards}  & 38.73 &  35.34 \\
AMT-GAN~\cite{hu2022protecting}  & 34.44  & 38.12\\
\midrule
\rowcolor{orange!6} Ours & 26.62 & 50.18 \\
\bottomrule[0.1em]
\end{tabular}}
\caption{\small FID comparison. PSR Gain is absolute gain in PSR relative to Adv-Makeup.}
\end{SCtable}

\subsection{Experimental Results} \label{sec:results}
In this section, we present experimental results of our approach in \textit{black-box} settings on four different pretrained \texttt{FR} models under face verification and identification tasks. 
To generate protected images, we use three \texttt{FR} models as a surrogate to imitate the decision boundary of the fourth \texttt{FR} model.
All results are averaged over 5 text based makeup styles that are provided in the supplementary material.

%%%%%%%%%%%%%%%verification%%%%%%%%%%%%%%
For face verification experiments, we set the system threshold value at 0.01 false match rate for each \texttt{FR} model \ie IRSE50 (0.241), IR152 (0.167), FaceNet (0.409), and MobileFace (0.302).
Quantitative results in terms of PSR for impersonation attack under the face verification task are shown in Tab. \ref{table:verification_impersonation}. 
Our approach is able to achieve an average absolute gain of about $12\%$ and $14\%$ over SOTA unrestricted~\cite{hu2022protecting} and noise-based~\cite{yang2021towards} facial privacy protection methods, respectively. Qualitative results are shown in Fig. \ref{fig:title_fig} which shows that protected faces generated by our approach are more realistic. Results for dodging attacks under face verification are provided in the supplementary material.
%%%%%%%%%%%%%%%%%%%%%%%%%%%%%%%%%%%%%%%%%%%%%%%%%%
In Tab. \ref{table:identification}, we also provide PSR vales under the face identification task for dodging (untargeted) and impersonation attacks. Our approach consistently outperforms recent methods at both \textit{Rank-1} and \textit{Rank-5} settings. We emphasize that we are the first to show effectiveness of generative models in offering untargeted privacy protection (dodging) in a more practical identification setting. Since AMT-GAN and Adv-Makeup are originally trained to impersonate target identity under the verification task, we have not included them in Tab. \ref{table:identification}. Qualitative results for LFW and CelebA are provided in the supplementary material.
%%%%%%%%%%%%%%%%%%%%%%%%%%%%%%%%%%%%%%%%%%%%%%%%%%

We report FID scores (lower is better) of our approach in Tab. \ref{table:fid} for CelebA and LADN datasets to measure naturalness. Adv-Makeup has the lowest FID score as it only applies makeup to the eye region without changing the rest of the face. However, this kind of restriction results in poor PSR. Our method has lower FID compared to TIP-IM and AMT-GAN and achieves the highest PSR. We provide PSNR and SSIM results in the supplementary material.

%Due to space limitations, we defer the Results comparison between using global clip loss (Eq. \ref{eq:global})  and directional clip loss (Eq. \ref{eq:dir}) are provided in the supplementary material.

\iffalse
\begin{table}[]
\centering
\caption{Evaluation metrics}
\resizebox{0.12\textwidth}{!}{ 
\begin{tabular}{|l|l|}
\hline
\textbf{Metrics}   & \textbf{}                   \\ \hline
ASR          &                     \\ \hline
FID &   \\ \hline
PSNR       &  \\ \hline
SSIM                &         \\ \hline
User study              &         \\ \hline
\end{tabular}}
\end{table}
\fi
%\vspace{-1m}
\subsection{Effectiveness in Real-World Applications} \label{sec:real-world}

\begin{figure}[t] 
\centering
\includegraphics[clip, trim=0cm 0.2cm 0cm 0cm,width=0.49\textwidth]{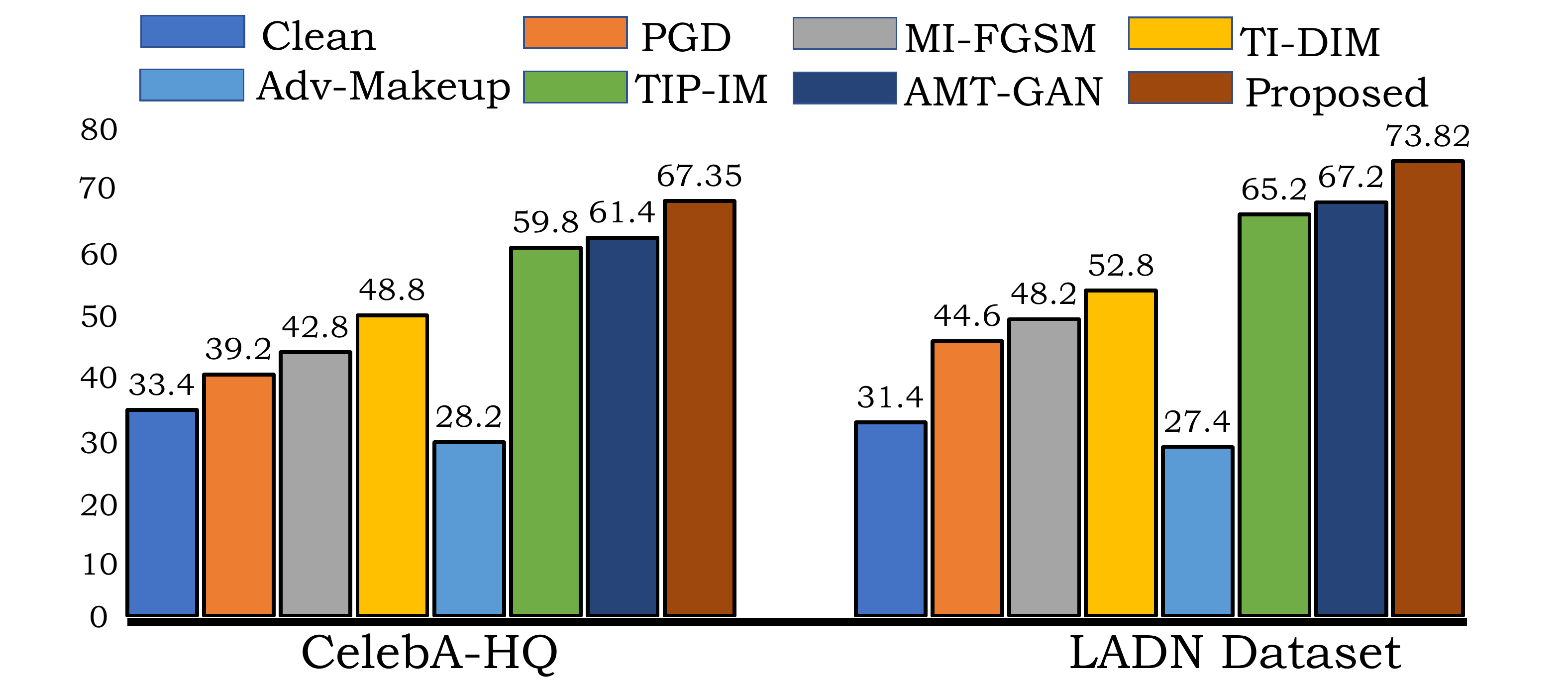}
\vspace{-1em}
\caption{Average confidence score (higher is better) returned by a real-world face verification API, Face++, for impersonation attack. Our approach has a higher confidence score than state-of-the-art makeup and noise-based facial privacy protection methods. }
\label{fig:face+}
\end{figure}

We now show the effectiveness of our approach to protect facial images (through targeted impersonation) against commercial API such as Face++ and Tencent Yunshentu \texttt{FR} platform operating in the verification mode. These APIs return confidence scores between $0$ to $100$ to measure whether two images are similar or not, where a high confidence score indicates high similarity. As the training data and model parameters of these propriety \texttt{FR} models are unknown, it effectively mimics a real-world scenario. We protect 100 faces randomly selected from CelebA-HQ using the baselines and the proposed method. 
In Fig. \ref{fig:face+}, we show the average confidence score returned by Face++ against these images. These results indicate that our method has a high PSR compared to baselines. We defer more details and results for Tencent Yunshentu API to supplementary material.

%We test the proposed approach in face verification setting. 

\label{subsec: Preliminaries}

\subsection{Ablation Studies} \label{sec:ablations}
Next, we report some ablations to evaluate the contributions of our loss components.

\begin{figure}[t] 
\centering
\includegraphics[width=0.36\textwidth]{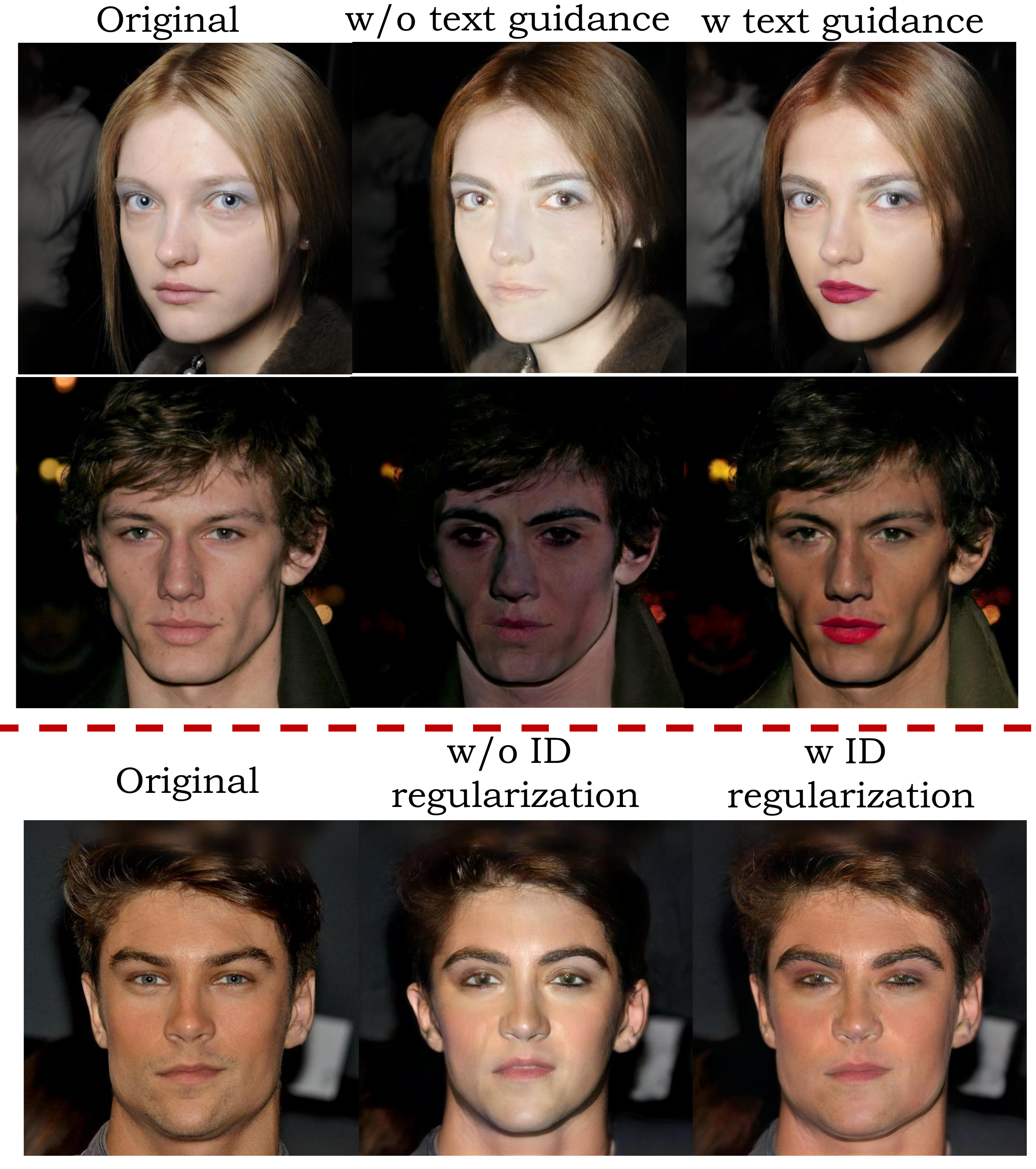}
\caption{Top: Effect of makeup-based text guidance on the visual quality of the output images.   Output images are able to impersonate the target identity for face verification. Text-prompt is \textit{``tanned makeup with red lipstick"}.  Bottom: Optimizing over all latent codes changes the identity of the protected image. Our identity-preserving regularization enforces the adversarial optimization to search for latent codes that hide the perturbations in the makeup effect while simultaneously preserving visual identity.}
\label{fig:woclip}
\end{figure}

\noindent \textbf{Makeup based text guidance}: As shown in Fig. \ref{fig:woclip} (top), in the absence of text guidance, resulting images may contain artifacts due to increased perturbations induced by the adversarial objective. Text-guidance effectively hides the perturbations in the makeup, leading to more natural looking images. It also provides the user more flexibility to select a desired makeup style compared to a reference image. 

\noindent \textbf{Identity preserving regularization}: Optimizing over the whole latent space provides more degrees of freedom and increases the PSR. However, it does not explicitly enforce adversarial optimization to preserve the user identity as shown in Fig. \ref{fig:woclip} (bottom). The proposed identity preserving regularization effectively preserves identity, while imitating the desire makeup style.

%%%%%%%%%%%%%%%%%%%%%%%%%%%%%%%%%%%%%%%%%%%%%%%%%%%%%5
%%%%%%%%%%%%%%%%%%%%%%%%latent loss %%%%%%%%%%%%%%%%%%
%%%%%%%%%%%%%%%%%%%%%%%%%%%%%%%%%%%%%%%%%%%%%%%%%%%%%%%%%%%%%%
\begin{table}[t]
\begin{center}
\caption{\small Impact of $\lambda_{\text{latent}}$ on FID score and PSR.}
\label{table:quant}
\vspace{-2mm}
\setlength{\tabcolsep}{7.0pt}
\scalebox{0.78}{
\begin{tabular}{l | c  | c  | c |c |c|c|c}
\toprule[0.15em]
\rowcolor{mygray} \textbf{$\lambda_{\text{latent}}$}&0.5 &0.1& 0.05 & 0.01 & 0.005 & 0.0001 & 0 \\
\midrule[0.15em]
 FID  &  11.6 & 21.4  & 25.2 &27.8  & 30.1& 38.4 &43.2\\
 PSR (\%)  &  31.2 & 39.0  & 57.4 & 76.2 & 83.8& 90.0 &93.6\\
\bottomrule[0.1em]
\end{tabular}}
\end{center}\vspace{-1.5em}
\end{table}

\noindent \textbf{Impact of latent loss weight}: Decreasing the weight assigned to the latent loss $\lambda_{\text{latent}}$ results in an increase in both the FID score and PSR (and vice versa). Allowing the latent to deviate more from the initial inverted latent code of the given face image often results in artifacts caused by the adversarial loss, degrading naturalness but aiding privacy.
%%%%%%%%%%%%%%%%%%%%%%%%%%%%%%%%%%%%%%%%%%%%%%%%%%%%%5
%%%%%%%%%%%%%%%%%%%%%%%%latent loss %%%%%%%%%%%%%%%%%%
%%%%%%%%%%%%%%%%%%%%%%%%%%%%%%%%%%%%%%%%%%%%%%%%%%%%%%%%%%%%%%
\begin{table}[t]
\begin{center}
\caption{\small Impact of different textual makeup styles on PSR. Makeup styles are \textit{``tanned"}, \textit{``pale"}, \textit{``pink eyeshadows"}, \textit{``red lipstick"}, and \textit{``Matte"}. Std. denotes standard deviation.}
\label{table:promt}
\vspace{-2mm}
\setlength{\tabcolsep}{7.0pt}
\scalebox{0.82}{
\begin{tabular}{l | c  | c  | c |c |c|c}
\toprule[0.15em]
\rowcolor{mygray} & $t_{\text{makeup}}^1$ &$t_{\text{makeup}}^2$& $t_{\text{makeup}}^3$ & $t_{\text{makeup}}^4$ & $t_{\text{makeup}}^5$ & Std.  \\
\midrule[0.15em]
PSR  &  74.1 & 77.3  & 78.4 & 78.7 & 79.2& 1.24 \\
\bottomrule[0.1em]
\end{tabular}}
\end{center}\vspace{-1.5em}
\end{table}

\noindent \textbf{Robustness against textual variations.} Finally, we evaluate the impact of different textual styles on the PSR. We select five text-based makeup styles to protect 1000 images of CelebA-HQ using our method. Results in Tab. \ref{table:promt} shows that PSR does not change significantly (low standard deviation) for different makeup styles, indicating robustness of our approach \emph{wrt} different text-based makeup styles.

%For all ablations, we provide more qualitative results in the {\color{red}supplementary material}.
\section{Conclusion}

We have proposed a framework to protect privacy of face images on online platforms by carefully searching for adversarial codes in the low-dimensional latent manifold of a pre-trained generative model. We have shown that incorporating a makeup text-guided loss and an identity preserving regularization effectively hides the adversarial perturbations in the makeup style, provides images with high quality, and preserves human-perceived identity. While this approach is robust to the user-defined text-prompt and target identity, it would be beneficial if the text-prompt and target identity can be automatically selected based on the given face image. Limitations of our method include high computational cost at the time of protected face generation.

\iffalse
\subsection{Ablation Studies}

\begin{itemize}
\item Table-1 will be of blackbox attack success rate on two datasets. CelebA and LADN. Comparison methods will be recent makeup based attack approaches including AMT-GAN (CVPR,2022) and Adv-makeup (IJCAI,2021) and noise based approaches including DIM, PGD. Depending on time, we can also add Adv-Hat, Adv-Eyeglasses, and Adv-Patch in comparison.
\item Figure-1 will be qualitative figure, where we compare our method with AMT-GAN, Adv-Makeup, and noise based approaches.
\item Table-2 will be of PSNR, SSIM, and FID score of output image with the original source image. Probably the results of our approach will be less as compared to other methods in terms of PSNR and SSIM.
\item Figure-2: Some bar plots for attack on online systems.
\item Figure-3: Experiments with optimizing over all layers of StyleGAN2 vs optimizing over last identity preserving layers. Attack success rate will be higher for all layers but identity will not be preserved in that case.
\item Table-3: Experiments by optimizing over both style and noise vectors of StyleGAN. 
\item Impact of prompts on the attack success rate.
\item Number of iterations experiments
\item Robustness against JPEG compression
\item May be experiments against robust FR models
\end{itemize}
\fi

% Use \bibliography{yourbibfile} instead or the References section will not appear in your paper
%%%%%%%%% REFERENCES
{\small
\bibliographystyle{ieee_fullname}
\bibliography{egbib.bib}
}

\end{document}